\documentclass[conference]{IEEEtran}
\IEEEoverridecommandlockouts
\pdfoutput=1 
\usepackage{cite}
\usepackage{amsmath,amssymb,amsfonts}
\usepackage{algorithmic}
\usepackage{longtable} 
\usepackage{graphicx}
\usepackage{textcomp}
\usepackage{array}
\usepackage{caption}
\usepackage{xcolor}
\usepackage{subfig}
\usepackage{url}            
\usepackage{booktabs}       
\usepackage{amsfonts}       
\usepackage{nicefrac}       
\usepackage{microtype}      
\def\BibTeX{{\rm B\kern-.05em{\sc i\kern-.025em b}\kern-.08em
    T\kern-.1667em\lower.7ex\hbox{E}\kern-.125emX}}
\begin{document}

\title{Use of Winsome Robots for Understanding Human Feedback (UWU)\\

}

\author{
\IEEEauthorblockN{Jessica Eggers}
\IEEEauthorblockA{\textit{Georgia Institute of Technology} \\
Atlanta, GA, USA \\
jeggers7@gatech.edu}
\and
\IEEEauthorblockN{Angela Dai}
\IEEEauthorblockA{\textit{Georgia Institute of Technology} \\
Atlanta, GA, USA\\
adai24@gatech.edu}
\and
\IEEEauthorblockN{Matthew C. Gombolay}
\IEEEauthorblockA{\textit{Georgia Institute of Technology}\\
Atlanta, Georgia, USA\\
matthew.gombolay@cc.gatech.edu}
}

\maketitle

\begin{abstract}
 As social robots become more common, many have adopted cute aesthetics aiming to enhance user comfort and acceptance. However, the effect of this aesthetic choice on human feedback in reinforcement learning scenarios remains unclear. Previous research has shown that humans tend to give more positive than negative feedback, which can cause failure to reach optimal robot behavior. We hypothesize that this positive bias may be exacerbated by the robot's level of perceived cuteness. To investigate, we conducted a user study where participants critique a robot's trajectories while it performs a task. We then analyzed the impact of the robot’s aesthetic cuteness on the type of participant feedback. Our results suggest that there is a shift in the ratio of positive to negative feedback when perceived cuteness changes. In light of this, we experiment with a stochastic version of TAMER which adapts based on the user's level of positive feedback bias to mitigate these effects.
\end{abstract}

\section{Introduction}
Human-robot interaction (HRI) is becoming more common as robots become integrated into everyday society. As robots become more widely adopted, inevitably they will encounter scenarios outside of their training scope and programmed capabilities. However, it is unrealistic to expect these robots to be configured for all new situations; therefore, to function successfully they must possess the ability to learn and adapt to new circumstances.

One promising approach for this challenge is Learning from Demonstration (LfD), a method where users teach robots tasks through demonstrations or by providing feedback on their performance. Unlike traditional programming approaches, LfD allows users to directly teach the robot without requiring technical expertise. This accessibility makes LfD particularly attractive for enabling general users to customize robot behavior and improve their performance. 

Such a high level of interaction necessitates that users feel comfortable and are willing to engage closely with the robot. To promote user comfort, recently many companies have released robots that have utilized ``cute" aesthetics, using features that align with the baby schema principle (a psychological schema based on infant features) proposed by Lorenz \cite{bcute, b9, b11}. Research suggests that robots whose appearance adheres to this cuteness schema were considered more trustworthy than their counterparts \cite{b11}. However, cuteness alone is insufficient to incite this engagement; users may still experience frustration if the robot lacks adequate skills or cannot perform the task\cite{b9}. Allowing users the ability to refine the robot's skills through training methods like LfD may help mitigate such frustrations.

\begin{figure}[t]
  \centering
      \includegraphics[width=\columnwidth ]{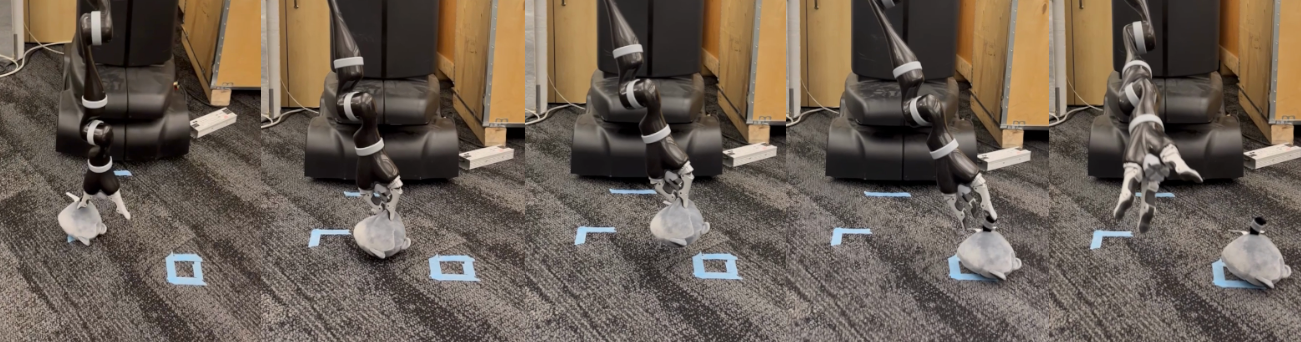}
      \caption{This figure depicts a robotic pick-and-place task. We record positive and negative feedback provided by users on the robot's task execution. Our paper explores the impact of robotic aesthetics, such as perceived cuteness, on users’ feedback tendencies and subsequent robot learning outcomes.}
      
\end{figure}

Many HRI and LfD studies use machine-like robots for their research, overlooking the potential impact of the robot's visual aesthetic on user feedback. In cases such as reinforcement learning where human feedback is the driver for shaping the robot’s learning process, it is important to understand how this aesthetic change may impact the type and quality of the user’s demonstrations or feedback. Prior HRI research has shown that humans tend to give more positive than negative feedback during training \cite{b4,b5,b6}. This positive bias can pose an issue to robotic learning algorithms as users may give positive feedback for poor robot performance, creating a noisy reward \cite{b4,b5}. Furthermore, the inaccurate positive feedback may cause the robot to enter a ``positive reward circuit," where it will become stuck in a loop of suboptimal states and never reach the goal \cite{b4}. Additionally, others have shown that appearance can influence a user’s rating of a robot’s trustworthiness and acceptance of a robot \cite{b7}. However, no previous study has directly investigated the relationship between the robot’s perceived cuteness and the type of feedback a user gives. This research aims to provide a framework for investigating this relationship between visual robotic cuteness and the quality of user feedback in an LfD setting.

In this work, we make the following contributions: 
\begin{itemize}
\item We conduct a proof-of-concept, IRB-approved (Protocol H24178), pilot study that investigates the impact of perceived robotic cuteness on user feedback.
\item We propose additions to Interactive Robot Learning (IRL) algorithm, TAMER, aimed at mitigating suboptimal positive feedback. 
\item We outline a framework for designing user studies to evaluate the impacts of cuteness on IRL algorithms. 
\end{itemize}

\section{Background}

Learning from demonstration (Lfd) is a diverse field that seeks to enable robots to be easily trained by end-users to perform assistive tasks \cite{b14, b15, b16, b17, b18, b19, b20}. Presently, there are open challenges, such as addressing specific nuances and biases in data provided by humans teaching robots \cite{b21, b22, b23, b24, b25}. We review TAMER, an Lfd algorithm that utilizes feedback from the user and the resulting feedback bias that can occur when suboptimal feedback is provided. We also explore the impacts of robot appearance and how it relates to positive feedback bias.

\subsection{TAMER}
TAMER or Training an Agent Manually via Evaluative Reinforcement is a reinforcement learning framework that allows a human expert to shape an agent or robot’s policy through feedback \cite{tamer}. The human expert observes the behavior of the agent and provides evaluative feedback as the agent performs its trajectory.  The TAMER agent models the human’s reinforcement and chooses actions that it expects to be the most reinforced. Relying on human feedback allows the agent to function without an environmental reward function and makes robot learning more accessible to people with less robotic knowledge. There are many LfD algorithms that utilize similar feedback methods such as Deep Tamer and COACH \cite{b26,b27,b28}. These algorithms have been incorporated into many reinforcement learning from human feedback (RLHF) systems \cite{b29, b30, b31, b32}.

\subsection{Positive Feedback Bias}
Previous testing with TAMER found that participants overall gave more positive than negative feedback during the training process. This tendency towards positive feedback increased with time as the robot’s performance improved \cite{b5}. Knox and Stone found that some participants would even give positive feedback despite extremely poor robot performance. In a task where the participant had to critique a TAMER agent’s performance on Tetris, 17\% of participants continued to give the agent positive feedback despite its terrible performance \cite{b4}. Knox and Stone further emphasize that this positive feedback can greatly impact the agent’s performance. Too much positive feedback may introduce positive reward circuits where an MDP agent can become stuck in a loop due to a circuit with an infinite expected reward, causing them to never reach the goal. In the future works section, we experiment with adapting this algorithm to be more robust against positive feedback bias. 

\subsection{Cuteness in Robots}
Current research explores how the appearance of a robot can impact its perceived sociability, but few studies address the impact of “cuteness” on feedback optimality. Findings indicate that tailoring a robot’s aesthetic can increase acceptance and comfort among participants who interact with it \cite{b8}. With regards to robotic cuteness, studies also show that robots that rate high on the baby schema are viewed as cuter and also more trustworthy than those that rank lower \cite{b11}.

Additionally, various studies have found that the appearance of a robot can affect how humans choose to interact with it. One study noted that participants who found a robot cute would use “baby talk” when addressing it \cite{b10}. Similarly, when presented with either a humanoid or quadruped robot, researchers found that participants would use human language such as “thank you” to address the humanoid robot, while using more pet-like language (ex: “good”) when addressing the quadruped robot \cite{quad}. This suggests that a robot’s appearance directly impacts the type of feedback given by humans. We could not find any work that directly linked a robot’s level of perceived cuteness to changes in the type or quality of human feedback.

We hypothesize that this positive feedback bias will be exacerbated if the users perceive the robot as cute. To analyze this effect, we performed a within-subject user study to investigate the impacts of cuteness on user feedback.  Furthermore, we experiment with a stochastic version of TAMER that breaks them out of these positive circuits to account for this feedback distribution shift. By adapting to the user's feedback bias, stochastic TAMER should enable an agent to learn more efficiently and accurately.

\section{Methods}
\subsection{Hypothesis}
 We aim to study the effect of perceived robot cuteness on participant feedback.
 \subsubsection{Perceived Robotic Cuteness}
 We first hypothesize that participants will find the robot with the “cute” face attached to be significantly cuter than the control. Previous research has shown that robots and inanimate objects which adhere stronger to the baby schema are perceived as ``cuter" by subjects \cite{b11}. We designed the cute face in accordance with the baby schema using bigger eyes lower down the face. For the control face, we used smaller eyes higher on the face. By utilizing the baby schema, we hypothesize that subjects will find the robot with a stronger adherence to the baby schema robot to be cuter than the control.
 
 \subsubsection{Cuteness Impact on Feedback}If the first hypothesis holds, we further hypothesize that the ratio of positive to negative feedback will be higher for the cute robot when compared to the control. In other words, participants will give the cute robot more positive feedback. Studies have shown that ``cuteness" can illicit strong feelings of protectiveness and positive feelings from the viewer towards the subject \cite{b9, caretaking}. Furthermore, users experienced similar affections towards human subjects as dogs, cats and even inanimate objects \cite{b13}. We hypothesize that these positive feelings may translate into an increase in positive feedback for the ``cute" robot and may exacerbate the positive feedback bias found in previous research such as that of Knox and Stone.

\subsection{Metrics}
We describe the metrics collected during the trials and through our pre-task and post-task surveys. First, we ask the user to fill out an initial survey to measure their general opinion on robots and teaching. For each robot (control and cute), we ask the users to complete a pre-task survey after they first see the robot but before feedback is given, and a post-task survey after the feedback session is complete (See Appendix for survey details).

\subsubsection{Overall User Perception of the Robot} Both pre-task and post-task, we use a 5 point Likert scale to measure the users overall opinion of the robot. We also have the users provide a short explanation for their rating. 
\subsubsection{User Perception of Robot Cuteness} Both pre-task and post-task, we use a 5 point Likert scale to measure the perceived cuteness of the robot. We have the users write a short comment explaining their rating.
\subsubsection{Feedback (Type \& Count)} During each trajectory for each robot, we collect the users binary feedback [negative, positive] for the robot's performance. 

\begin{table}[h!]
\centering
\renewcommand{\arraystretch}{1.2} 
\begin{tabular}{|p{3cm}|p{4.5cm}|}
\hline
\textbf{Question} & \textbf{Answer Options} \\
\hline
1. How are you feeling today? & 
\begin{tabular}[t]{@{}l@{}}
1: Negative emotions \\ 
2: Semi-negative emotions \\ 
3: Neutral \\ 
4: Semi-positive emotions \\ 
5: Positive emotions 
\end{tabular} \\
\hline
2. In general, how do you feel about robots? &
\begin{tabular}[t]{@{}l@{}}
1: Negative sentiments \\ 
2: Semi-negative sentiments \\ 
3: Neutral \\ 
4: Semi-positive sentiments \\ 
5: Positive sentiments 
\end{tabular} \\
\hline
3. How comfortable do you feel about working with robots? & 
\begin{tabular}[t]{@{}l@{}}
1: Not at all comfortable \\ 
2: Uncomfortable \\ 
3: Neutral \\ 
4: Comfortable \\ 
5: Very comfortable 
\end{tabular} \\
\hline
4. What is your previous experience with teaching robots? &
\begin{tabular}[t]{@{}l@{}}
1: No experience \\ 
2: Limited experience \\ 
3: Moderate experience \\ 
4: Extensive experience 
\end{tabular} \\
\hline
5. What is your prior experience with robots in general? &
\begin{tabular}[t]{@{}l@{}}
1: No experience \\ 
2: Limited experience \\ 
3: Moderate experience \\ 
4: Extensive experience 
\end{tabular} \\
\hline
6. How do you feel about teaching others? &
\begin{tabular}[t]{@{}l@{}}
1: Strongly dislikes teaching others \\ 
2: Dislikes teaching others \\ 
3: Neutral about teaching others \\ 
4: Likes teaching others \\ 
5: Strongly likes teaching others 
\end{tabular} \\
\hline
\end{tabular}
\caption{Pre-Experiment Survey}
\label{table:pre_experiment_survey}
\end{table}

\begin{table}[h!]
\centering
\renewcommand{\arraystretch}{1.2} 
\begin{tabular}{|p{3cm}|p{4.5cm}|}
\hline
\textbf{Question} & \textbf{Answer Options} \\
\hline
1. How do you feel about this robot in general? & 
\begin{tabular}[t]{@{}l@{}}
1: Negative sentiments \\ 
2: Semi-negative sentiments \\ 
3: Neutral \\ 
4: Semi-positive sentiments \\ 
5: Positive sentiments 
\end{tabular} \\
\hline
2. Explain your rating (open-ended) & 
[Open-ended response] \\
\hline
3. Rank the robot’s cuteness on a scale of 1-5 & 
\begin{tabular}[t]{@{}l@{}}
1: Not at all cute \\ 
2: Not cute \\ 
3: Neutral \\ 
4: Cute \\ 
5: Very cute 
\end{tabular} \\
\hline
4. Can you elaborate on your rating? & [Open-ended response] \\
\hline
5. How comfortable do you feel teaching this robot? & 
\begin{tabular}[t]{@{}l@{}}
1: Very uncomfortable \\ 
2: Uncomfortable \\ 
3: Neutral \\ 
4: Comfortable \\ 
5: Very comfortable 
\end{tabular} \\
\hline
\end{tabular}
\caption{Pre-Task Survey}
\label{table:post_experiment_survey}
\end{table}
\begin{table}[h!]
\centering
\renewcommand{\arraystretch}{1.2} 
\begin{tabular}{|p{3cm}|p{4.8cm}|}
\hline
\textbf{Question} & \textbf{Answer Options} \\
\hline
1. How has your opinion of the robot changed after this session? & 
\begin{tabular}[t]{@{}l@{}}
1: Increased negative sentiments by a lot \\ 
2: Increased negative sentiments by a little \\ 
3: The same opinion \\ 
4: Increased positive sentiments by a little \\ 
5: Increased positive sentiments by a lot 
\end{tabular} \\
\hline
2. Please explain your rating & [Open-ended response] \\
\hline
3. Rerank the robot’s cuteness on a scale of 1-5 & 
\begin{tabular}[t]{@{}l@{}}
1: Not at all cute \\ 
2: Not cute \\ 
3: Neutral \\ 
4: Cute \\ 
5: Very cute 
\end{tabular} \\
\hline
4. How comfortable do you feel about this robot after this session? & 
\begin{tabular}[t]{@{}l@{}}
1: Very uncomfortable \\ 
2: Uncomfortable \\ 
3: Neutral \\ 
4: Comfortable \\ 
5: Very comfortable 
\end{tabular} \\
\hline
\end{tabular}
\caption{Post-Task Survey}
\label{table:post_session_survey}
\end{table}

\subsection{Pilot Study Experimental Design}
To evaluate the effect of perceived cuteness on user feedback, we conducted a within-subject user study with 14 participants. Each participant was tasked with critiquing a set of four pre-recorded pick-and-place task trajectories for both the control and cute robot. 
We prerecorded 12 different trajectories using the MOVO robot's Jaco arm for a simple pick-and-place task illustrated in Figure 1. 
\begin{figure}[h]
  \centering
  \includegraphics[width=5cm]{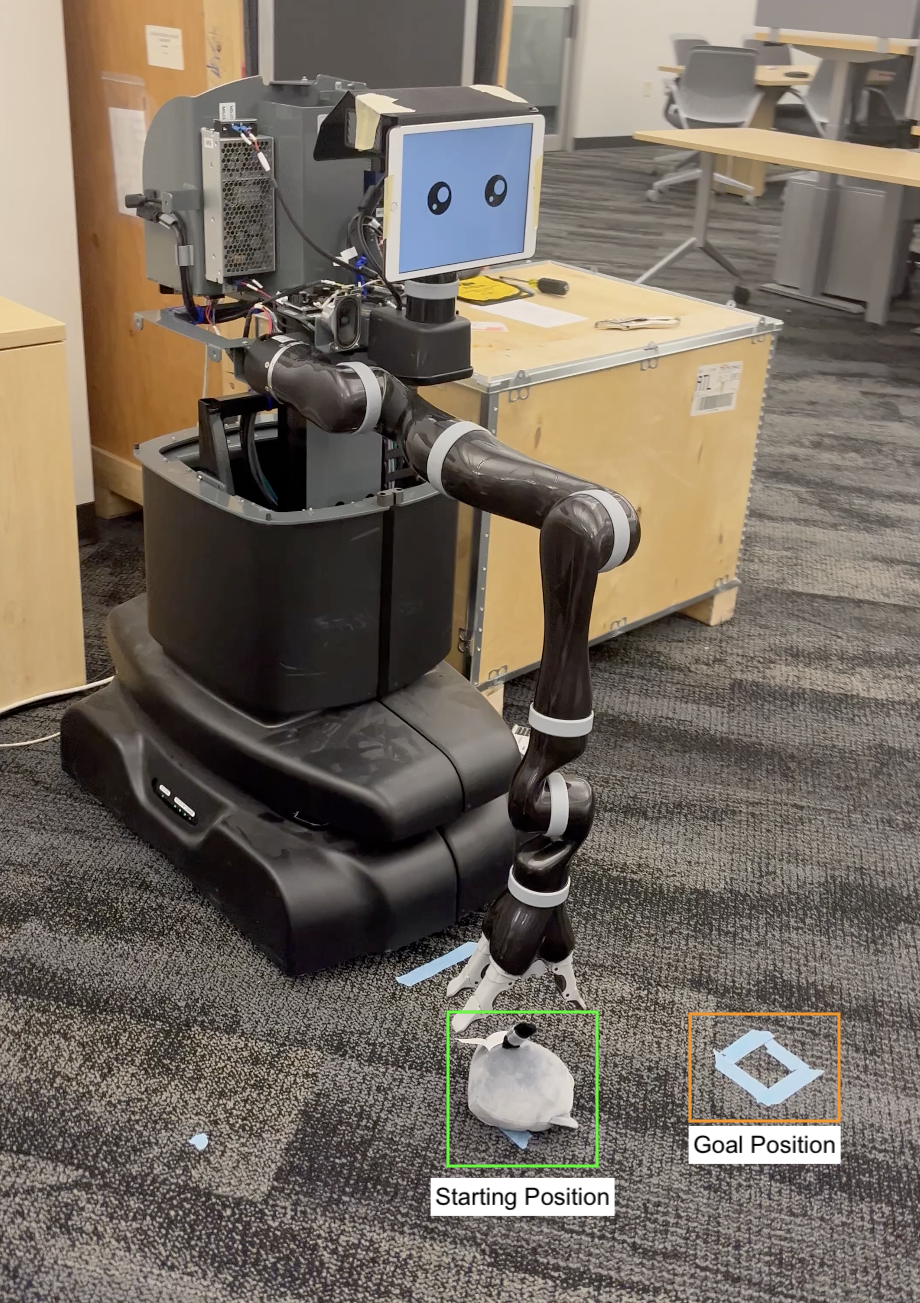}
  \caption{Illustration of the robot’s task. The MOVO must pick up the plushie from the starting position (current placement) and place it into the goal.}
\end{figure}

 Half of the trajectories were “successes” where the MOVO was either able to accomplish the task or very close to reaching the success state. The other half of the trajectories were “failures” which consisted of poor performance where the MOVO unmistakably did not complete the goal.
 
\subsubsection{Robot Design}
The MOVO robot by Kinova Labs was used as the base of both the control and cute robots. The robot performed the pick and place task with its 7 DOF Jaco arm. To change the face between trials, we attached an iPad to the front of the MOVO’s Kinetic sensor, which we then used to display the two faces as seen in Figure 2.
The cute and control faces were created in adherence to the baby schema \cite{b11}. For the cute robot, we wanted it to have a high baby schema, so we used larger eyes in the middle of the face to increase the forehead size. We also included highlights on the eyes to increase perceived eye size. For the control robot, we reduced the schema by placing the eyes farther up the face and reducing their size. To reduce confounding factors we kept the shape and color of the eyes the same between both trials. Also, we did not include additional features on the face to limit extraneous variables.

\begin{figure}[h]
  \centering
  \includegraphics[width=\columnwidth]{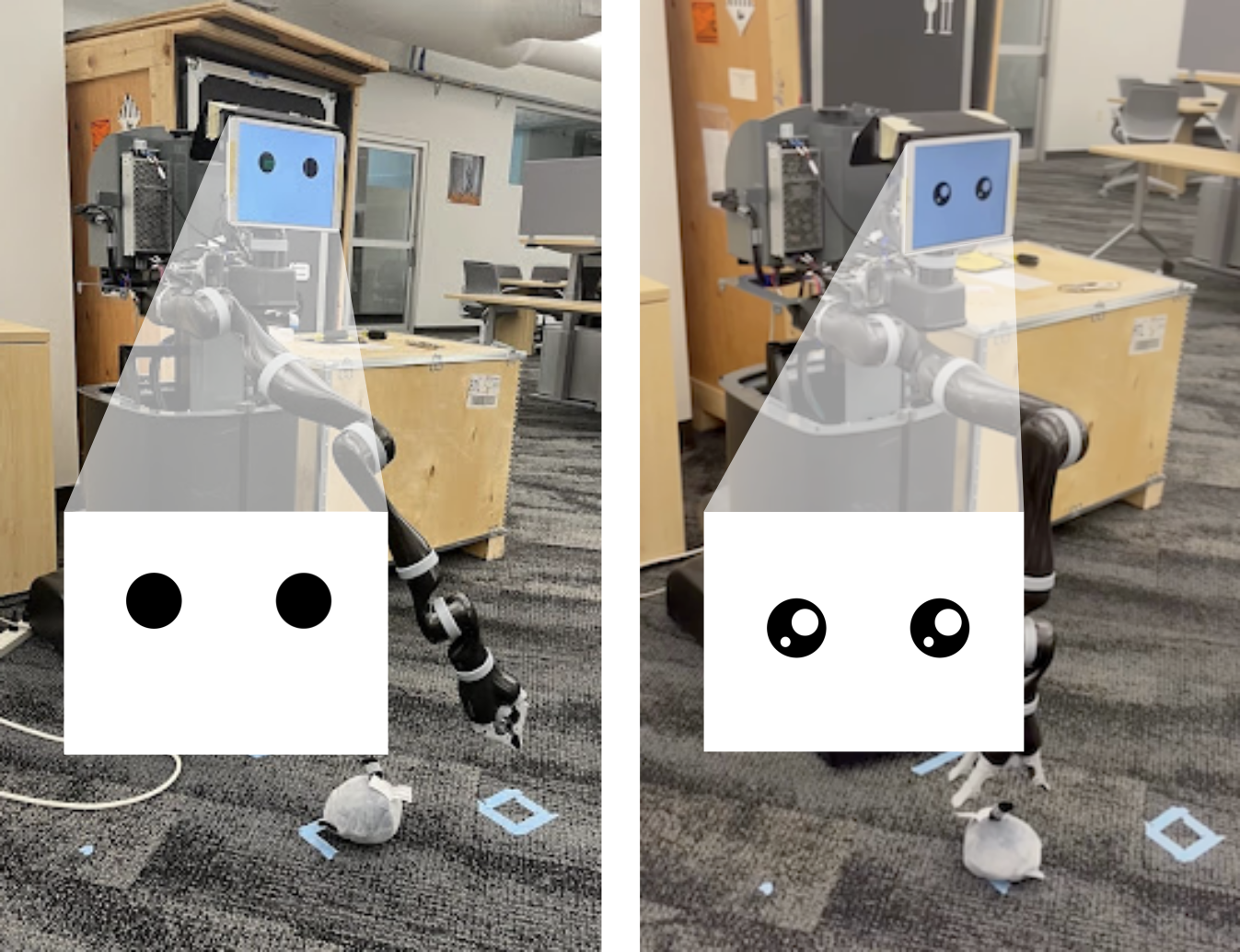}
  \caption{Robots faces used in the study. Control on the left and Cute on the right. The cute robot has a face with the appearance of larger eyes and forehead, features derived from the baby schema \cite{b11}.}
\end{figure}

\subsubsection{Participant Task}
Participants first completed a pre-task survey which involved completing a consent form for the study using DocuSign. This survey also collected data regarding general sentiments about robots and teaching. Then, after being showed the robot, participants completed another survey which asked users how cute they found the robot and was asked to explain their rating.
Each participant then received a random sample of four pre-recorded trajectories. These trajectories contained two trajectories where the robot was able to complete the goal during the recording period and two trajectories where the robot performed extraneous actions or incorrectly performed the task. 
During each trajectory playback, participants had access to a laptop with which they could press “n”  or “p” on the keyboard to give the robot negative (“n”) or positive (“p”) critiques throughout the trajectory. These feedback values were saved and parsed during later analysis. 
Before and after critiquing each robot, participants took a pre-training and post-training survey to record the perceived cuteness level of the robot and other qualitative data.

\subsection{Algorithm Development}
Having found potential evidence for positive bias in human feedback due to cuteness, we aimed to update TAMER to mitigate the subsequent negative learning effects \cite{b12}.

Our implementation adds a user score that evaluates the optimality of the user’s feedback. This score evaluates if the  user gave appropriate feedback for an action, where the appropriateness is determined by the environmental reward for that action. Thus, the user score is penalized when users give positive feedback to non-optimal actions. Then, if the user score dips below a pre-set threshold, the algorithm will stochastically swap positive feedback for negative with probability $p$. If the user score continues to be below the threshold, this probability will gradually increase causing greater randomness for consistent suboptimal positive feedback. 

\section{Experimental Results and Discussion}

\subsection{Perceived Cuteness of the Robot}
To verify that there was a difference in perceived cuteness between the two faces, we conducted a paired t-test on the survey results where $t(13) = 3.19, p < 0.01$. Participants were asked to rate the cuteness of the robot, giving it a score from 1-5, 1 being not cute at all and 5 being very cute. The mean perceived cuteness for the cute robot was 4.0  $\pm$ 0.877 out of 5, while the control robot received a 2.79 $\pm$ 1.25 out of 5. There were some outliers among participants, but overall most participants found the cute robot to be significantly cuter than the control. 

Qualitatively, participants found the cute robot to be more friendly than the control. In the survey data, when asked to explain their rating participants used words like ``friendly” and ``approachable” to describe the cute robot. The response sentiment for the control robot varied more. Participants described the appearance of the control robot as ``dumb,” ``no feeling” and ``cool”. For the control, multiple participants reported the robot having a “blank stare” and being ``unsettling”; however others reported the round eyes were still ``cute”. Furthermore, sentiment analysis shows that on average participants had a more positive initial response to the ``cute" robot when asked ``How do you feel about this robot in general?" We labeled the responses to this question on a scale of [-1, 0, 1], -1 indicating the participant had a negative sentiment in their response, and 1 indicating positive sentiment. Labels for responses to the cute robot averaged 0.615 $\pm  0.457$ for the cute robot. In contrast, the participants responses only scored 0.036 $\pm$ 0.930 for the control robot.

Both the quantitative scores and qualitative responses support our hypothesis H1, suggesting that the robot with the ``cute" face adhering to the baby schema was perceived as significantly cuter by the users than the control one.

Interestingly, in both scenarios, some participants distinguished the cuteness of the robot's face from the rest of the robot, providing one cuteness explanation for the face and one for the body. For example, ``Eyes are cute but everything else is normal." Additionally, many participants seemed to empathize with both robots, giving them encouragement during training or saying the robot ``is trying his best."

\subsection{Cuteness impact on Feedback}
We found that participants gave a greater ratio of positive to negative feedback for the cute vs control robot ($t(14) = 1.66, p < 0.1$) for a paired one-tailed T-test. Furthermore, participants generally gave much more positive than negative feedback: mean positive/negative feedback: cute = 2.65 $\pm$ 2.26, control = 1.71 $\pm$ 1.23. Since the data suggests that participants found the ``cute" robot to be cuter, the robot’s cuteness is a likely motivator for this feedback shift.

\begin{figure}[h]
  \centering
  \includegraphics[width = \columnwidth]{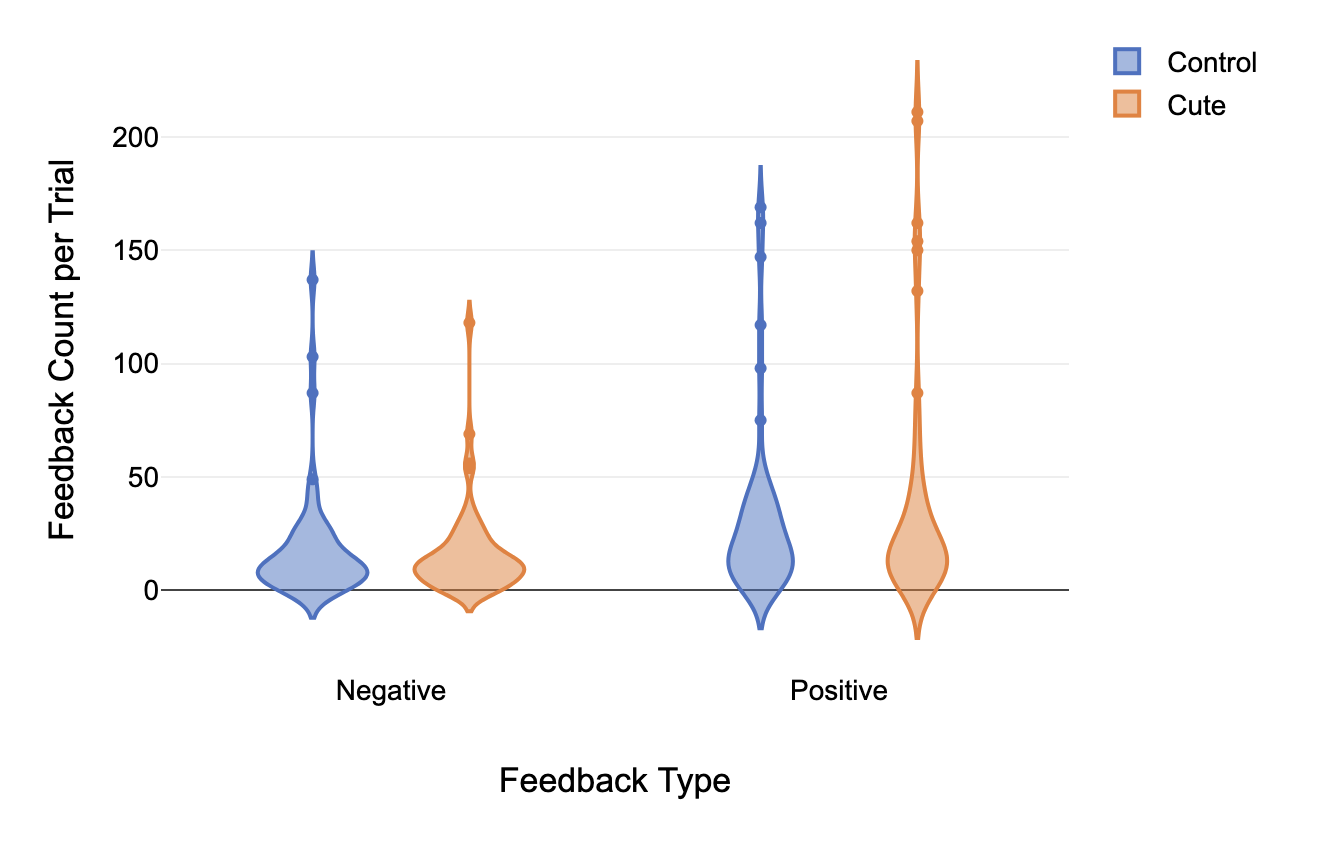}
  \caption{Comparison of the distribution of positive and negative feedback for the two robots. Users were less likely to give negative feedback for the cute than the control (left), and overall gave more positive feedback to the cute robot than the control (right).}
\end{figure}

Due to our small sample size and other limiting factors, we were not able to obtain a p-value below the conventional threshold of 0.05 for our feedback ratio analysis ($t(14) =
1.66$, $p = 0.061$). However, these results indicate a pattern trending towards significance  suggesting a potential relationship between perceived cuteness and feedback ratio. While we cannot confidently state that our second hypothesis (H2) is supported, the observed trend, where users appeared to provide more positive feedback to the robot perceived as cuter, warrants further exploration in future studies.

\subsection{Impact of Feedback Type on Reinforcement Learning}
\begin{figure}[h]
  \centering
  \includegraphics[width = 0.5\columnwidth]{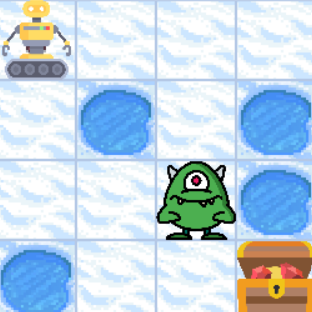}
  \caption{Wumpus World with robot agent (top left). Obstacles include holes and a monster. Goal state is the treasure box (bottom right).}
  \label{fig:ww}
\end{figure}

\begin{figure}[h]
  \centering
  \includegraphics[width = 1\columnwidth, height = 6cm, keepaspectratio, trim={1mm 1mm 1mm 1mm},clip]{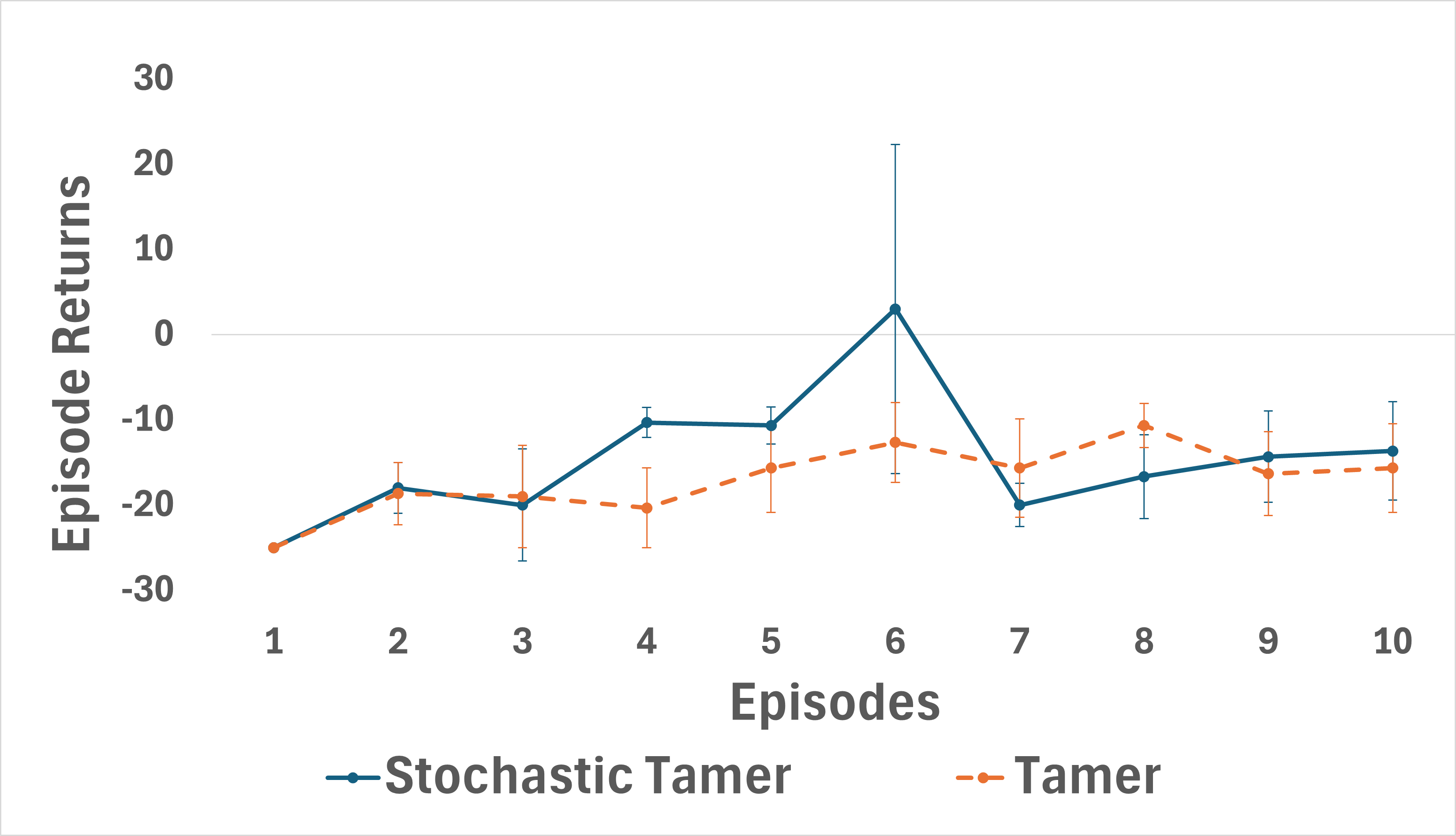}
  \caption{Comparison of Stochastic TAMER vs TAMER episode returns with 50\% positive feedback.}
  \label{fig:50}
\end{figure}

\begin{figure}[h]
  \centering
  \includegraphics[width = 1\columnwidth, height = 6cm, keepaspectratio, trim={1mm 1mm 1mm 1mm},clip]{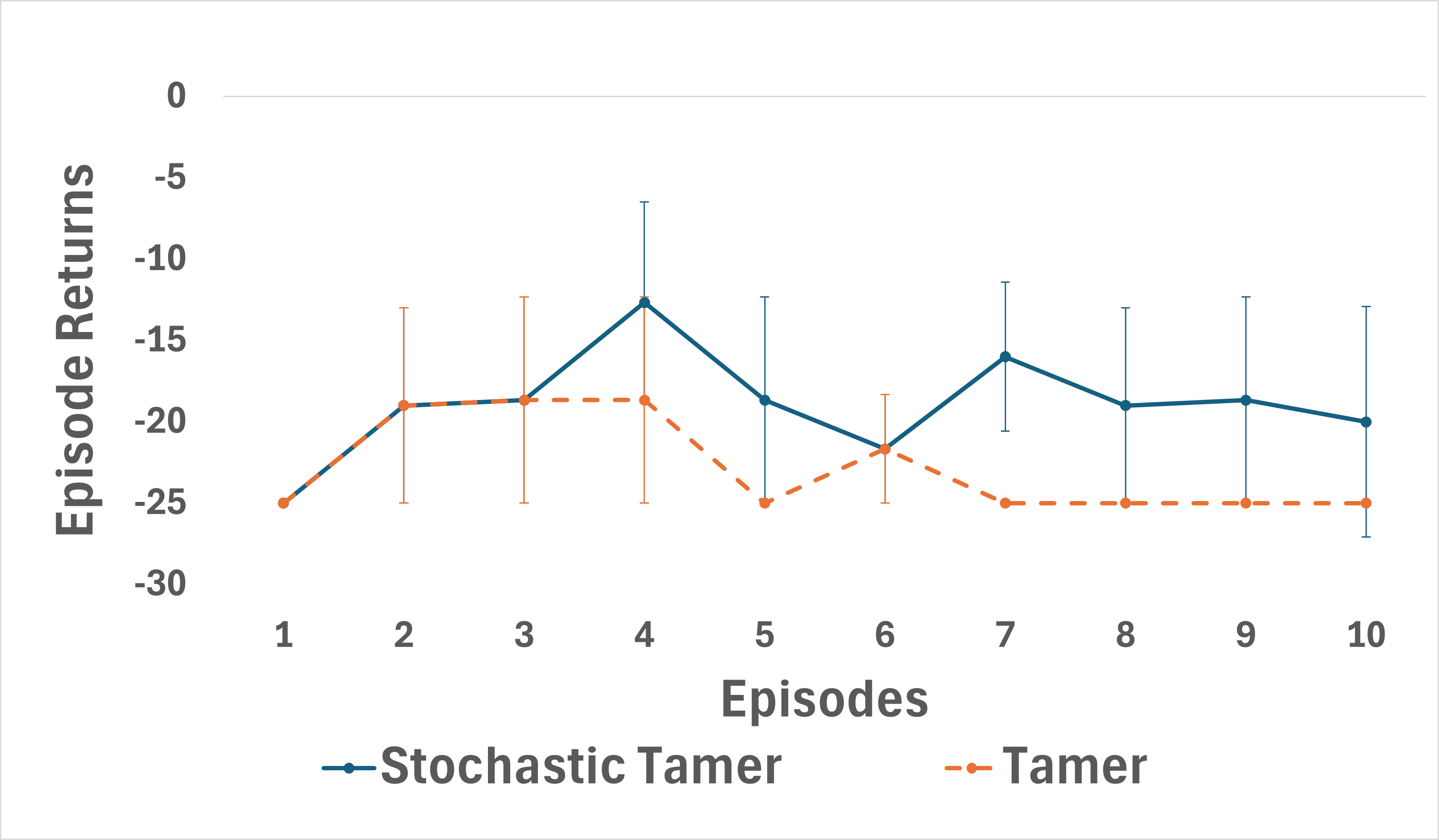}
  \caption{Comparison of Stochastic TAMER vs TAMER episode returns with 66\% positive feedback.}
  \label{fig:60}
\end{figure}

\begin{figure}[h]
  \centering
  \includegraphics[width = 1\columnwidth, trim={1mm 1mm 1mm 1mm},clip]{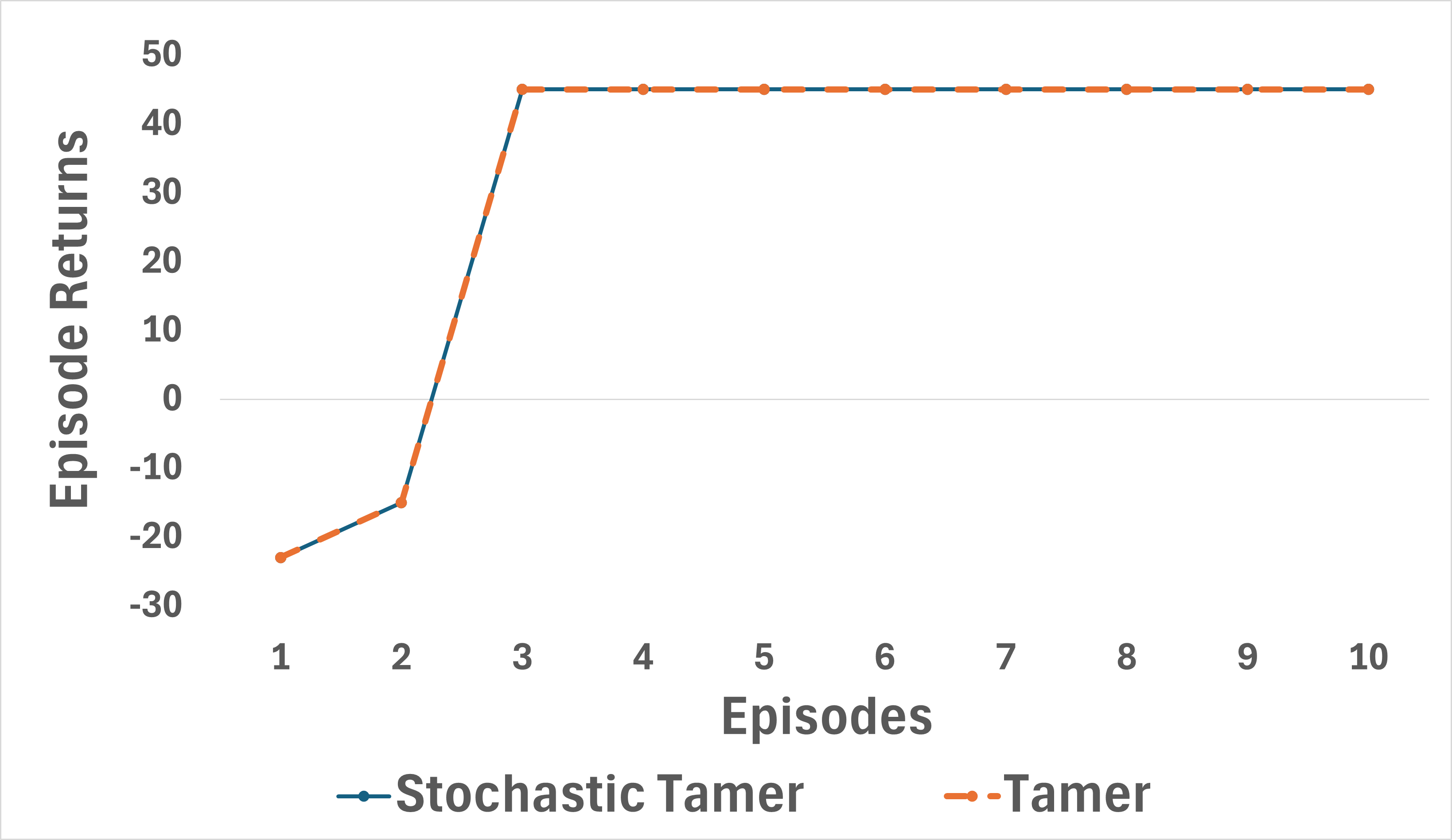}
  \caption{Stochastic TAMER and TAMER episode returns with optimal feedback.}
  \label{fig:stochasticoptimal}
\end{figure}

We added modifications to the TAMER algorithm as explained in Methods, Part D, to create Stochastic TAMER. Our updated algorithm allows the robot to break from a positive reward circuit by randomly including negative feedback. We tested the algorithm in Wumpus World (Figure \ref{fig:ww}), where the robot moves in cardinal directions. After each move, the user provides either positive or negative feedback. We conducted three trials and gave three different levels of positive feedback. As seen in Figure \ref{fig:50}, Stochastic TAMER when using 50\% positive feedback had higher overall episode returns and saw positive returns in the sixth episode. Additionally, tests with 66.67\% positive feedback, seen in Figure \ref{fig:60}, showed that TAMER was unable to achieve higher returns than -18. On the other hand, the updated algorithm was able to achieve higher returns, albeit it did not break 0 within 10 episodes. For optimally given feedback, this algorithm exhibited the same environmental returns as the original TAMER for each episode, as seen in Figure \ref{fig:stochasticoptimal}. This finding was consistent across each testing session, leading to a standard deviation of zero. Due to time constraints, implementation and testing were limited. More investigation is needed to create a more robust algorithm to counteract positive bias. This investigation indicates that robots who receive more positive feedback will have a harder time learning how to reach the goal state without intervention from the algorithm. 

As evidenced by the previous sections, humans can perceive robots to be ``cuter'' due to aesthetic shifts. Additionally, there appears to be a potential trend where users provide more positive feedback to the ``cuter" robot. These factors combined with the results above suggest that a more winsome robot may receive positively skewed feedback inadvertently causing poor performance with conventional learning algorithms not designed to handle the skewed data.

\section{Limitations and Future Work}

There were also some potential confounding factors. One factor being the order in which participants saw the two robots. On average, participants gave the first robot they saw a higher overall rating than the second (3.86 $\pm$ 0.86 vs 3.35 $\pm$ 1.44). Furthermore, participants who saw the cute robot second gave it a higher cuteness rating than those who saw it first (first = 3.71 $\pm$ 0.95, second = 4.29 $\pm$ 0.76). Similarly those who saw the normal robot first gave it a higher cuteness rating than those who saw it second (first = 3.14 $\pm$ 1.07, second = 2.43 $\pm$ 1.40). We hypothesize that users may have been using their previous experience as a baseline when providing their second rating. Additionally, due to time and resource constraints, each participant only had a very short window to complete both rounds of critiques. With extra time, participants could participate in a wash-out task between the two critique sessions; however, due to the within-subject nature of the study, there will most likely still be some persistence bias.

We also had some complications with the MOVO robot itself. The robot’s motion planner is stochastic, adding an element of unpredictability, so there were some initial challenges with recording usable and consistent trajectories. Furthermore, on the second day of the user study, the gripper attached to the JACO arm stopped functioning and suddenly started working again for the last three participants. Due to time pressure, we were unable to rerecord the participants' trajectories with the working gripper; however, although their overall sentiment toward the robot decreased, the participants faced with the broken gripper still displayed a similar pattern for the feedback ratio as the working-gripper participants.

\section{Design Takeaways}
We gathered key takeaways from our pilot study for future work investigating the impacts of cuteness. 

Isolating the cuteness features: In order to prevent confounding variables such as anthropomorphism, we elected to change only one feature of the robot - the eyes. Additional feature changes could increase human-likeness of one robot over the other which research has shown impacts user perception \cite{b12}.

Choosing reliable hardware: We struggled to have consistent robot trajectories due to hardware malfunctions. This introduced factors that impacted the user's perception of the robot. For future study designs, we suggest utilizing reliable software and hardware to avoid this issue. 

User task: We had users give feedback to prerecorded robot trajectories. This provided consistency for each trajectory, hence we could analyze feedback across each trajectory type. We did not use a Wizard of Oz technique; users were not told that their feedback impacted the current trajectories, but rather were told that their feedback would be used for later processing. This prevented the possibility of inconsistency where a user gave non optimal feedback but the robot improved in the task. We also wanted to avoid any frustration that arose when the robot did not follow a user's feedback which could impact the user's perception of the robot.

\section{Conclusion}
Prior work has shown that perceived cuteness in robots can impact a user's comfort and trust of the agent. We extend this research by focusing on the impact of perceived cuteness on feedback in a reinforcement scenario. Our study indicates that perceived cuteness does have a statistical impact on a human's ratio of positive to negative feedback, with the cute robot receiving a higher ratio of positive to negative feedback than the control. Additionally, we experiment with a stochastic version of TAMER to mitigate potential issues stemming from the positive feedback bias. To do so we inject negative feedback when the user is deemed too suboptimal by the algorithm. Our study demonstrates the significance of perceived cuteness on user feedback, highlighting the need for adaptive approaches like  stochastic TAMER to address potential bias and enhance the effectiveness of human-robot interaction in reinforcement scenarios.

\section{Acknowledgments}
We thank the teaching staff of CS 7648: Interactive Robot Learning for providing the source code for the base TAMER algorithm. We also thank the CORE lab for granting access to the MOVO robot used in this study. Finally, we acknowledge Arthur Scaquetti for his valuable assistance with the robot.

\end{document}